%% file: paper.tex
\documentclass[a4paper,10pt]{article}
\usepackage[utf8]{inputenc}

\usepackage{algorithmicx}
\usepackage[noend]{algpseudocode}

\usepackage[linguistics]{forest}
\usepackage{amsfonts}
\usepackage{amsmath}
\usepackage{enumitem}  

\usepackage{cite}

\usepackage{geometry}[margin=1in]

\usepackage{supertabular}

\usepackage{multicol}
\usepackage{hyperref} 

\usepackage{textcomp}
\usepackage{listings}

\definecolor{listinggray}{gray}{0.9} 
\definecolor{lbcolor}{rgb}{0.98,0.98,0.98}
\usepackage{xcolor} \definecolor{Darkgreen}{rgb}{0,0.4,0}
\newcommand{\pair}[2]{\left\langle #1 , #2 \right\rangle}
\title{How to enumerate trees\\from a context-free grammar}
\author{Steven T. Piantadosi}

\begin{document}

\maketitle

\begin{abstract}
I present a simple algorithm for enumerating the trees generated by a Context Free Grammar (CFG). The algorithm uses a pairing function to form a bijection between CFG derivations and natural numbers, so that trees can be uniquely decoded from counting. This provides a general way to number expressions in natural logical languages, and potentially can be extended to other combinatorial problems. I also show how this algorithm may be generalized to more general forms of derivation, including analogs of Lempel-Ziv coding on trees. 
\end{abstract}

\section{Introduction}

While context-free grammars (CFGs) are important in computational linguistics and theoretical computer science, there is no simple, memoryless algorithm for enumerating the trees generated by an arbitrary CFG. One approach is to maintain a priority queue of partially expanded trees according to probability, and expand them through (e.g.) the leftmost unexpanded nonterminal in the tree. This, however, requires storing multiple trees in memory, which can become slow when enumerating many trees. Incremental polynomial time algorithms are also known \cite{costa2015naive} and related questions have been studied for lexicographic enumeration \cite{makinen1997lexicographic,domosi2000unusual,dong2009linear}. These algorithms are not particularly well-known, and the tools required to state and analyze them are complex. In contrast, simple techniques exist for enumerating binary trees  with a fixed grammar (e.g. $S \to S S \mid x $). A variety of techniques and history is reviewed in Section 7.2.1.6 of \cite{knuth2006art_trees}, including permutation-based methods and gray codes \cite{semba1981generation,skarbek1988generating,zaks1980lexicographic,er1985enumerating}. These algorithms, however, do not obviously generalize to arbitrary CFGs. 

The goal of the present paper is to present an variant of integer-based enumeration schemes that works for arbitrary CFGs. The algorithm is itself very basic---just a few lines---but relies on a abstraction here called an \lstinline{IntegerizedStack} that may be useful in other combinatorial problems. The proposed algorithm does not naturally enumerate in lexicographic order (though variants may exist) but it is efficient: its time complexity is linear in the number of nodes present in the next enumerated tree, and it does \emph{not} require additional data structures or pre-computation of anything from the grammar. Because the algorithm constructs a simple bijection between a the natural numbers $\mathbb{N}$ and trees, it also provides a convenient scheme for G\"{o}del-numbering \cite{nagel2012godel,smullyan1992godel}, when the CFG is used to describe formulas. We then extend this algorithm to a tree-based algorithms analogous to LZ compression.

\section{Pairing functions}

\begin{figure} 
\centering
 \hspace{25pt}
 \includegraphics[scale=0.8]{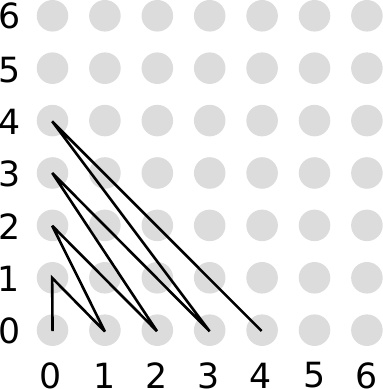} 
  \hspace{25pt}
 \includegraphics[scale=0.8]{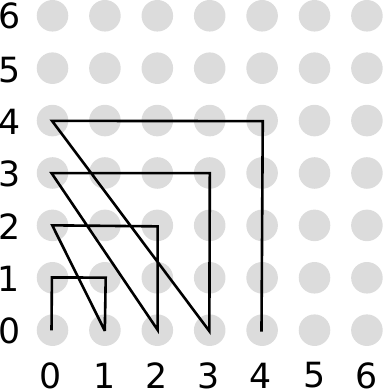} 
  \hspace{25pt}
 \includegraphics[scale=0.8]{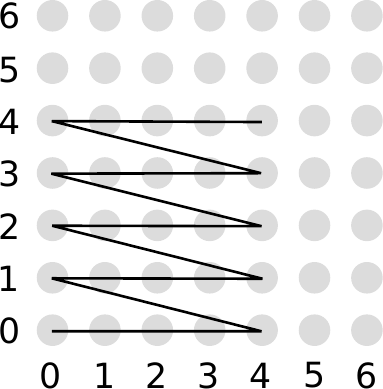}
 \caption{ Enumeration order of Cantor's pairing function (left), the Rosenberg-Strong pairing function (center) and the $M_4(x,y)$ (right).} 
 \label{fig:pairing}
\end{figure}

To construct a bijection between trees and integers, we use a construction that has its roots in Cantor \cite{cantor1878beitrag}'s proof that the rationals can be put into one-to-one correspondence with the integers. Cantor used a \emph{pairing function} \cite{szudzik2017rosenberg} to match up $\mathbb{N}\times\mathbb{N}$ with $\mathbb{N}$ itself:
\begin{equation}
C(x,y) = \frac{(x+y)\cdot(x+y+1)}{2} + y
\end{equation}
This function essentially traces the position of an integer pair $\pair{x}{y}$ in the line shown in Figure \ref{fig:pairing}. This pairing function is (uniquely) invertible via
\begin{equation}
\pair{x}{y} = C^{-1}(z) =  \pair{z-\frac{w\cdot(w+1)}{2}}{\frac{w\cdot(w+3)}{2}-z},
\end{equation}
for $w = \lfloor \frac{1}{2}(-1+\sqrt{1+8z}) \rfloor$. This function has, interestingly, been the study of additional formal work. It is, for example, the only quadratic bijection between $\mathbb{N} \times \mathbb{N}$ and $\mathbb{N}$ \cite{vsemirnov2001two,adriaans2018simple}; an analysis of the computational complexity of different pairing functions can be found in \cite{regan1992minimum}.

Other pairing functions are more convenient for some applications. A popular alternative, illustrated in Figure \ref{fig:pairing} is the Rosenberg-Strong pairing function \cite{rosenberg1972addressing}, 
\begin{equation}
 R(x,y) = max(x,y)^2 + max(x,y) + x - y
\end{equation}
with inverse,
\begin{equation}
R^{-1}(z) 
=
\begin{cases}
\pair{z-m^2}{m}          & \text{ if }  z-m^2 < m \\
\pair{m}{m^2 + 2m - z}   & \text{otherwise,}
\end{cases}
\end{equation}
where $m=\lfloor \sqrt{z} \rfloor$. Pairing functions are reviewed in \cite{szudzik2017rosenberg}, who also shows how they may be used to enumerate binary trees. The key idea is that we can imagine that any integer $n$ is a pairing of its two subtrees (e.g. $n=R(x,y)+1$ for subtrees $x$ and $y$). If we iterate over integers, we may then ``translate'' each integer into a binary tree by breaking it down into two integers and then recursively doing the same on $x$ and $y$ until we reach $0$. Specifically, assume that 
\begin{equation} \label{eq:localphi}
 \phi(R(x,y)+1) = \pair{\phi(x)}{\phi(y)}.
\end{equation}
Then, for example, $n=147$ can be broken down as,
\begin{equation}
\begin{aligned}
& 147 \\
& \pair{2}{12} \\
& \pair{\pair{0}{1} }{\pair{2}{3}} \\
& \pair{\pair{0}{\pair{0}{0}} }{\pair{2}{3}} \\
& \pair{\pair{0}{\pair{0}{0}} }{\pair{\pair{0}{1}}{\pair{1}{1}}} \\
& \pair{\pair{0}{\pair{0}{0}} }{\pair{\pair{0}{\pair{0}{0}}}{\pair{\pair{0}{0}}{\pair{0}{0}}}}. \\
& \pair{\pair{\bullet}{\pair{\bullet}{\bullet}} }{\pair{\pair{\bullet}{\pair{\bullet}{\bullet}}}{\pair{\pair{\bullet}{\bullet}}{\pair{\bullet}{\bullet}}}}. \\
\end{aligned}
\end{equation}
So long as $R$ is any max-dominating\footnote{A function $f$ such that $f(x,y)>max(x,y)$.} pairing function, $\phi$ is an enumeration of trees \cite{szudzik2017rosenberg}.  

It may not also be obvious how to use this approach to generate from an arbitrary CFG where the productions allowed at each step vary depending on the CFG's nonterminal types. In particular, there may be multiple ways of expanding each nonterminal, which differs depending on which non-terminal is used. A simple scheme such as giving each CFG rule an integer code and then using a pairing function like $R$ to recursively pair them together will not, in general, produce a bijection because there may be integer codes that do not map onto full trees (for instance, pairings of two terminal rules in the CFG). 

The issue is that in generating from a CFG, we have to encode a choice of which rule to expand next, of which there are only finitely many options. In fact, the number of choices will in general depend on the nonterminal. Our approach to address this is to use two different pairing functions: a modular ``pairing function'' to encode which nonterminal to use and the Rosenberg-Strong pairing function to encode integers for the child of any node. Thus, if a given nonterminal has $k$ expansions, define a pairing function that pairs $\lbrace 0,1,2,\ldots,k-1 \rbrace \times \mathbb{N}$ with $\mathbb{N}$. A simple mod operation, shown in Figure \ref{fig:pairing}, will work: 
\begin{equation}
 M_k(x,y) = x + k\cdot y
\end{equation}
with inverse
\begin{equation}
 M_k^{-1}(z) = \left\langle z \mod k, \frac{z-(z \mod k)}{k} \right\rangle.
\end{equation}

\section{Enumerating trees}

It is convenient to combine mod pairing and Rosenberg-Strong pairing into a simple abstraction, we here call an  \lstinline{IntegerizedStack}. This term is intentionally different from ``integer stack'' (which is a stack of integers): an \lstinline{IntegerizedStack} is a stack of integers that is itself \emph{stored in an integer}. This class allows us to pack and unpack a finite list of integers from a single integer, using \lstinline{push} and \lstinline{pop} operations of a standard stack. For use later, we can push or pop raw integer, or modulo some number:
\begin{figure}
\begin{lstlisting}[language=python]
class IntegerizedStack:
    def __init__(self, v=0):
        self.value = v

    def pop(self):
        # remove an integer from self.value and return
        self.value, ret = decode(self.value)
        return ret

    def modpop(self, modulus):
        # pop from self.value mod n
        self.value, ret = mod_decode(self.value, modulus)
        return ret

    def split(self, n):
        # Assume value codes exactly n integers. Zero afterwards.
        out = [self.pop() for _ in range(n-1)]
        out.append(self.value)
        self.value = 0
        return out
\end{lstlisting}
\end{figure}
Here, we have assumed that \lstinline{decode} is the inverse of a pairing function like $R^{-1}$ above, and \lstinline{mod_decode} is $M_k^{-1}$. Note, that the stored value of an \lstinline{IntegerizedStack} is always only an integer, but the abstraction of an \lstinline{IntegerizedStack} allows us to treat it as though it currently contains a stack of other integers, either through the pairing function or through a modulo pairing function. This stack has the special property that popping a stack with value $0$ always returns $0$ and leaves the stack with value $0$. \lstinline{IntegerizedStack} also includes one special helper function  \lstinline{split}, which partitions the integer into $k$ different components by successive \lstinline{pop}s. Note that this operation exhaustively uses up the remainder of the stack and leaves it empty. This could alternatively be achieved using a pairing function on $\mathbb{N}^d$ (see \cite{szudzik2017rosenberg}). 

To see how an \lstinline{IntegerizedStack} works in enumeration, let us assume we are working with a context-free grammar $\mathcal{G} = (V, \Sigma, R, S)$, where $V$ is a set of nonterminal symbols, $\Sigma$ is a set of terminal symbols, $R$ is a relation $V \to (V \cup \Sigma)^*$, and $S \in V$ is a start symbol. We require notation to distinguish the \emph{terminal} from \emph{non-terminal} rules in $R$. Let $T_v \subseteq R$ denote the set of \emph{terminal} rules, meaning those that expand $v$ with \emph{no} non-terminals on the right hand side (i.e. such that $v \to \Sigma^*$). Let $N_v \subset R$ be those that expand $v$ to some nonterminal. Following typical implementations, we will talk about $T_v$ and $N_v$ as an ordered list of rules. 

Without loss of generality, we make two further assumptions about $\mathcal{G}$:
\begin{enumerate}[label=(\roman*)]
 \item For each $v \in V$, the set of trees that $v$ can expand to is infinite. Note that any non-finite context-free language $\mathcal{G}$ can be converted into this format by, for instance, taking any $v \in V$ which only expands to finitely many trees, giving each tree a unique terminal symbol, and then removing $v$ from the grammar. This will create a new grammar $\mathcal{G}'$ whose productions can be translated back and forth to those of $\mathcal{G}$ with appropriate transformation of the new terminal symbols.
 \item The rule ordering in $G$ must be such that choosing the first (zeroth) rule for each nonterminal will eventually yield a terminal. This ensures that the first ($0$'th) item in any enumeration is a finite tree. 
\end{enumerate}
In practice, it will often be useful to put the terminals and then high-probability expansions \emph{first} in each $N_v$.
 
The generating algorithm, denoted \textbf{Algorithm A}, is then very simple. To expand an integer $n$ for nonterminal type $v \in V$, we first check if $n < |T_v|$ and if so, we return the $n$'th terminal rule. Otherwise, we treat $n-|T_v|$ as an \lstinline{IntegerizedStack}. We  \lstinline{pop} modulo $|N_v|$ to determine which rule expansion to follow, and then use \lstinline{split} to  specify the integers for the children, which are then recursively expanded. Algorithm A is shown in the function \lstinline{from_int} which takes a nonterminal type and an integer \lstinline{n}, and constructs a tree:
\begin{lstlisting}
# Given a nonterminal (string), an integer n, and cfg (a hash 
# from strings to lists of right hand side expansions), return 
# the n'th tree. Here, Node is a simple class that stores a 
# nonterminal Node.nt and a list of children (Nodes or strings), 
# Node.children. 
def from_int(nt, n, cfg):
    
    # count the nonterminals
    nterminals = sum([is_terminal_rule(rhs, cfg) for rhs in cfg[nt]])

    if n < nterminals: 
        # if n is coding a nonterminal
        return Node(nt, cfg[nt][n])
    else:
        # Treat n-nterminals as a stack of integers
        i = IntegerizedStack(n - nterminals)

        # how many nonterminal rules
        nnonterminals = len(cfg[nt]) - nterminals

        # i first encodes which *non*-terminal
        rhs = cfg[nt][nterminals+i.modpop(nnonterminals)]

        # split the remaining into the nonterminals on the right 
        # side of the rule
        t = i.split(sum( is_nonterminal(r, cfg) for r in rhs))

        # now we can expand all of the children
        children = []
        for r in rhs:
            if is_nonterminal(r,cfg):
                children.append(from_int(r,t.pop(0), cfg))
            else:
                children.append(Node(r))
                
        # Return the new Node
        return Node(nt, children)
\end{lstlisting}
In this listing, we have assumed that \lstinline{cfg} is a dictionary from nonterminal string symbols to lists of rules obeying (i) and (ii). In this algorithm, assumption (i) guarantees that any value of $n$ can be converted into a tree. Assumption (ii) ensures that when $n$ is zero, the algorithm will halt. Note that in both the mod and Rosenberg-Strong pairing functions, a value of $0$ will be unpaired into two zeros. This means that generally, at some point in the algorithm, the call to \lstinline{from_int} will take zero as an argument, and so (ii) is required to ensure that, in this case, it returns a finite tree rather than running forever. 

It may be counterintuitive in Algorithm A that we subtract $|T_v|$ from \lstinline{n}. This is required for \lstinline{from_int} to be a bijection, but the argument is more clear the inverse algorithm (converting trees to integers). If a tree with nonterminal $v \in V$ only consists only of a terminal rule, we simply specify which rule. Otherwise, we use an \lstinline{IntegerizedStack} to encode the nonterminal rule (modulo $|N_v|$) and all of the children. However, we do not want to give this \lstinline{IntegerizedStack} a number which overlaps with $0,1,\ldots,|T_v|-1$ since that would be confusable for a terminal rule. To avoid this, we start indexing the child trees at $|T_v|$. It should be clear, then, that this pairing is a bijection between trees and integers, for grammars satisfying (i) and (ii). 

An implementation of this algorithm is provided in the author's library github\footnote{Available at \url{https://github.com/piantado/enumerateCFG}} which is distributed under GPL. As an example, Figure \ref{fig:enum} shows expansions from a simple CFG that one might find in a natural language processing textbook:
\begin{equation} \label{eq:grammar}
 \begin{aligned}
   S  &\to NP\; VP \\
   NP &\to n \mid d\; n \mid d\; AP\; n \mid NP\; PP\\
   AP &\to a \mid a\; AP \\
   PP &\to p\; NP\\
   VP &\to v \mid v\; NP \mid v\; S \mid VP\; PP
 \end{aligned}
\end{equation}
   
\begin{figure} \tiny
\setlength{\columnsep}{0.5cm}
\begin{center}
\begin{multicols}{4} \raggedright
\input{strings.tex}
\end{multicols}
\end{center}
\caption{Enumeration of the grammar in (\ref{eq:grammar}) using Algorithm A.}
\label{fig:enum}
\end{figure}
        
Note that this encoding is a bijection between trees and integers, though not necessarily between terminals strings (yields) and integers, due to ambiguity in the grammar. Any number specifies a unique derivation, and vice-versa, giving rise to the bijection between trees and integers. The key assumption of this algorithm is context-freeness since that allows the pairings for each child of the tree to be independently expanded. However, a similar approach may be amenable to other combinatorial problems.

\section{LZ-trees}

An interesting family of variants to Algorithm A can be created by noting that an integer can encode information other than rule expansions in the grammar. For example, an integer might reference complete subtrees that have been generated previously. This idea is inspired by work formulating probabilistic models that expand CFGs to favor re-use of existing subtrees \cite{johnson2007adaptor,o2015productivity}. We call this approach LZ-trees because we draw on an idea from the LZ77/LZ78 algorithm \cite{ziv1977universal}, which compresses strings by permitting pointers back to previously emitted strings. Here, we permit our enumeration to potentially point back to previously generated complete trees. For instance, suppose we are currently decoding an integer at the point $x$ in the tree. 
\begin{equation}
\begin{forest}
 [A [B [C] [D [E] [F]]] [x] ]
\end{forest}
\end{equation}
Then at $x$, we should be allowed to draw on prior complete trees (rooted at $B$ and $D$) assuming they are of the correct non-terminal type for $x$. Since there are two previously-generated trees when expanding $x$, we can let \lstinline{IntegerizedStack} values of $0$ and $1$ reference these trees, and otherwise, we encode the node below $x$ according to Algorithm A. This has the effect of preferentially re-using subtrees that have previously been generated early in the enumeration, although it should be noted that the mapping is no longer a bijection. Note that unlike LZ77, this algorithm does not require us to store an integer for the length of the string/tree that is pointed to, because we assume it is a complete subtree. Also, the integer pointing to a previous tree is an integer specifying the target in any enumeration of nodes in the tree. A listing for this algorithm, Algorithm B, is shown below:

\begin{lstlisting}[language=Python]
# return a list of possible subtrees of t that LZ could reference
# usually we will want these to be complete subtrees involving more than
def possible_lz_targets(nt, T):
    out = []
    if T is not None:
        for t in T:
            if (t not in out) and (len(t) >= 3) and t.complete and t.nt == nt:
                out.append(t)
    return out

# provide the n'th expansion of nonterminal nt
def from_int(nt, n, cfg, root=None):

    # count up the number of terminals
    nterminals = sum([is_terminal_rule(rhs, cfg) for rhs in cfg[nt]])

    # How many trees could LZ reference?
    lz_targets = possible_lz_targets(nt, root)

    if n < len(lz_targets): 
        # we are coding an LZ target
        return deepcopy(lz_targets[n]) # must deepcopy
    elif n-len(lz_targets) < nterminals:
        # check if n is a terminal (remember to subtract len(lz_targets))
        return Node(nt, cfg[nt][n-len(lz_targets)])
    else:
        # n is what's leftover after trying to code lz_targets and terminals
        n = n - len(lz_targets) - nterminals

        # n-nterminals should be an IntegerizedStack where we
        i = IntegerizedStack(n)

        # how many nonterminal rules
        nnonterminals = len(cfg[nt]) - nterminals

        # i first encodoes which *non*-terminal
        which = i.modpop(nnonterminals)
        rhs = cfg[nt][nterminals+which]

        # count up how many on the rhs are nonterminals
        # and divide i into that many integers
        t = i.split(sum( is_nonterminal(r,cfg) for r in rhs))

        # A little subtlety: we have to store whether the node
        # is "complete" so we can know not to use it in recursive
        # calls until all its expansions are done
        out = Node(nt) # must build in children here
        out.complete = False
        for r in rhs:
            if is_nonterminal(r,cfg):
                out.children.append(from_int(r, t.pop(0), cfg, \\ 
                                    root if root is not None else out))
            else:
                # else it's just a string -- copy 
                out.children.append(r)
        # now the node is copmlete
        out.complete = True
        
        return out
\end{lstlisting}

Results from enumerating the grammar in (\ref{eq:grammar}) are shown in Figure \ref{fig:lzw}. Note here that the main differences are places where the Algorithm B re-uses a component earlier in the string. However, the algorithms do agree in many places, likely because of the requirement that only complete subtrees of the same type can be references (of which there are often not any). 
Similar approaches might allow us to write potentially write any kind of encoder and enumerate trees relative to that encoding scheme. For instance, we might permit a pointer to a previous \emph{subtree}, we might use an integer coding which codes prior tree components relative to their frequency, etc.

 \begin{figure} \label{fig:lzw} \tiny
\setlength{\columnsep}{1cm}
\begin{center}
\begin{multicols}{2} \raggedright
\input{strings-lzw.tex}
\end{multicols}
\end{center}
\caption{Enumeration of the grammar in (\ref{eq:grammar}) using Algorithm B. Lines only show strings where Algorithm A and Algorithm B give different answers.}
\end{figure}

\section{Conclusion}

This work describes a simple algorithm that enumerates the trees generated by a CFG by forming a bijection between these trees and integers. The key abstraction, an \lstinline{IntegerizedStack}, allowed us to encode arbitrary information into a single integer through the use of pairing functions.

\bibliographystyle{IEEEtran}
\bibliography{/home/piantado/Desktop/Science/Libraries/AllCitations,/home/piantado/Desktop/Science/Webpage/citations} 

\end{document}

%% file: strings.tex
0  nv \\
1  dnv \\
2  dnvn \\
3  nvn \\
4  danv \\
5  danvn \\
6  danvnv \\
7  dnvnv \\
8  nvnv \\
9  npnv \\
10  npnvn \\
11  npnvnv \\
12  npnvpn \\
13  danvpn \\
14  dnvpn \\
15  nvpn \\
16  daanv \\
17  daanvn \\
18  daanvnv \\
19  daanvpn \\
20  daanvdn \\
21  npnvdn \\
22  danvdn \\
23  dnvdn \\
24  nvdn \\
25  dnpnv \\
26  dnpnvn \\
27  dnpnvnv \\
28  dnpnvpn \\
29  dnpnvdn \\
30  dnpnvdnv \\
31  daanvdnv \\
32  npnvdnv \\
33  danvdnv \\
34  dnvdnv \\
35  nvdnv \\
36  daaanv \\
37  daaanvn \\
38  daaanvnv \\
39  daaanvpn \\
40  daaanvdn \\
41  daaanvdnv \\
42  daaanvnpn \\
43  dnpnvnpn \\
44  daanvnpn \\
45  npnvnpn \\
46  danvnpn \\
47  dnvnpn \\
48  nvnpn \\
49  dnpdnv \\
50  dnpdnvn \\
51  dnpdnvnv \\
52  dnpdnvpn \\
53  dnpdnvdn \\
54  dnpdnvdnv \\
55  dnpdnvnpn \\
56  dnpdnvdan \\
57  daaanvdan \\
58  dnpnvdan \\
59  daanvdan \\
60  npnvdan \\
61  danvdan \\
62  dnvdan \\
63  nvdan \\
64  daaaanv \\
65  daaaanvn \\
66  daaaanvnv \\
67  daaaanvpn \\
68  daaaanvdn \\
69  daaaanvdnv \\
70  daaaanvnpn \\
71  daaaanvdan \\
72  daaaanvdnvn \\
73  dnpdnvdnvn \\
74  daaanvdnvn \\
75  dnpnvdnvn \\
76  daanvdnvn \\
77  npnvdnvn \\
78  danvdnvn \\
79  dnvdnvn \\
80  nvdnvn \\
81  npdnv \\
82  npdnvn \\
83  npdnvnv \\
84  npdnvpn \\
85  npdnvdn \\
86  npdnvdnv \\
87  npdnvnpn \\
88  npdnvdan \\
89  npdnvdnvn \\
90  npdnvnpdn \\
91  daaaanvnpdn \\
92  dnpdnvnpdn \\
93  daaanvnpdn \\
94  dnpnvnpdn \\
95  daanvnpdn \\
96  npnvnpdn \\
97  danvnpdn \\
98  dnvnpdn \\
99  nvnpdn \\
100  daaaaanv \\

%% file: strings-lzw.tex
5 danvdan danvn \\
6 danvdanv danvnv \\
10 npnvnpn npnvn \\
11 npnvnpnv npnvnv \\
13 danvpdan danvpn \\
17 daanvdaan daanvn \\
18 daanvdaanv daanvnv \\
19 daanvpdaan daanvpn \\
20 daanvn daanvdn \\
21 npnvn npnvdn \\
22 danvn danvdn \\
26 dnpnvdnpn dnpnvn \\
27 dnpnvdnpnv dnpnvnv \\
29 dnpnvn dnpnvdn \\
30 dnpnvnv dnpnvdnv \\
31 daanvnv daanvdnv \\
32 npnvnv npnvdnv \\
33 danvnv danvdnv \\
37 daaanvdaaan daaanvn \\
38 daaanvdaaanv daaanvnv \\
39 daaanvpdaaan daaanvpn \\
40 daaanvn daaanvdn \\
41 daaanvnv daaanvdnv \\
42 daaanvdaaanpdaaan daaanvnpn \\
43 dnpnvdnpnpn dnpnvnpn \\
44 daanvdaanpdaan daanvnpn \\
45 npnvnpnpn npnvnpn \\
46 danvdanpdan danvnpn \\
50 dnpdnvdnpdn dnpdnvn \\
51 dnpdnvdnpdnv dnpdnvnv \\
52 dnpdnvpdn dnpdnvpn \\
53 dnpdnvn dnpdnvdn \\
54 dnpdnvnv dnpdnvdnv \\
55 dnpdnvdnpdnpdn dnpdnvnpn \\
56 dnpdnvdn dnpdnvdan \\
57 daaanvdn daaanvdan \\
58 dnpnvdn dnpnvdan \\
59 daanvdn daanvdan \\
60 npnvdn npnvdan \\
61 danvdn danvdan \\
65 daaaanvdaaaan daaaanvn \\
66 daaaanvdaaaanv daaaanvnv \\
67 daaaanvpdaaaan daaaanvpn \\
68 daaaanvn daaaanvdn \\
69 daaaanvnv daaaanvdnv \\
70 daaaanvdaaaanpdaaaan daaaanvnpn \\
71 daaaanvdn daaaanvdan \\
72 daaaanvnvdaaaan daaaanvdnvn \\
73 dnpdnvnvdnpdn dnpdnvdnvn \\
74 daaanvnvdaaan daaanvdnvn \\
75 dnpnvnvdnpn dnpnvdnvn \\
76 daanvnvdaan daanvdnvn \\
77 npnvnvnpn npnvdnvn \\
78 danvnvdan danvdnvn \\
82 npdnvnpdn npdnvn \\
83 npdnvnpdnv npdnvnv \\
84 npdnvpdn npdnvpn \\
85 npdnvn npdnvdn \\
86 npdnvnv npdnvdnv \\
87 npdnvnpdnpdn npdnvnpn \\
88 npdnvdn npdnvdan \\
89 npdnvnvnpdn npdnvdnvn \\
90 npdnvnpdnpnpdn npdnvnpdn \\
91 daaaanvdaaaanpn daaaanvnpdn \\
92 dnpdnvdnpdnpdnpdn dnpdnvnpdn \\
93 daaanvdaaanpn daaanvnpdn \\
94 dnpnvdnpnpdnpn dnpnvnpdn \\
95 daanvdaanpn daanvnpdn \\
96 npnvnpnpnpn npnvnpdn \\
97 danvdanpn danvnpdn \\
101 daaaaanvdaaaaan daaaaanvn \\
102 daaaaanvdaaaaanv daaaaanvnv \\
103 daaaaanvpdaaaaan daaaaanvpn \\
104 daaaaanvn daaaaanvdn \\
105 daaaaanvnv daaaaanvdnv \\
106 daaaaanvdaaaaanpdaaaaan daaaaanvnpn \\
107 daaaaanvdn daaaaanvdan \\
108 daaaaanvnvdaaaaan daaaaanvdnvn \\
109 daaaaanvdaaaaanpn daaaaanvnpdn \\
110 daaaaanvdaaaaan daaaaanvnpn \\
111 npdnvdan npdnvnpn \\
112 daaaanvdaaaan daaaanvnpn \\
113 dnpdnvdan dnpdnvnpn \\
114 daaanvdaaan daaanvnpn \\
115 dnpnvdan dnpnvnpn \\
116 daanvdaan daanvnpn \\
117 npnvdan npnvnpn \\
118 danvdan danvnpn \\
122 danpnvdanpn danpnvn \\
123 danpnvdanpnv danpnvnv \\
125 danpnvdan danpnvdn \\
126 danpnvdanv danpnvdnv \\
127 danpnvdanpnpn danpnvnpn \\
128 danpnvn danpnvdan \\
129 danpnvdanvdanpn danpnvdnvn \\
130 danpnvdanpnpdanpn danpnvnpdn \\
131 danpnvdn danpnvnpn \\
132 danpnvdanpnvdanpn danpnvnvn \\
133 daaaaanvdaaaaanvdaaaaan daaaaanvnvn \\